\documentclass[runningheads]{llncs}

\usepackage[mobile]{eccv}

\usepackage{eccvabbrv}

\usepackage{graphicx}
\usepackage{booktabs}
\usepackage[dvipsnames]{xcolor}

\usepackage{graphicx}
\usepackage{amsmath}
\usepackage{amssymb}
\usepackage{booktabs}
\usepackage{tabulary}
\usepackage{multirow}
\usepackage[accsupp]{axessibility}
\usepackage{forloop}
\usepackage{bm}

\newcommand{\tocite}[1]{\textcolor{red}{[TOCITE]}}
\newcommand{\toexp}[1]{\textcolor{blue}{[TOEXP]}}
\newcommand{\PAR}[1]{\noindent{{\textbf{#1~}}}}

\newcolumntype{x}[1]{>{\centering\arraybackslash}p{#1pt}}

\newcommand{\methodname}{Street Gaussians}
\newcommand{\methodnameblank}{\methodname\ }

\usepackage[accsupp]{axessibility}  

\usepackage[pagebackref,breaklinks,colorlinks]{hyperref}

\usepackage{orcidlink}

\begin{document}

\title{Street Gaussians: Modeling Dynamic Urban Scenes with Gaussian Splatting} 

\titlerunning{Street Gaussians for Modeling Dynamic Urban Scenes}


\author{
    Yunzhi Yan\inst{1, 2}
    \quad Haotong Lin\inst{1} 
    \quad Chenxu Zhou\inst{1} 
    \quad Weijie Wang\inst{1}  \\
    \quad Haiyang Sun\inst{2}  
    \quad Kun Zhan\inst{2} 
    \quad Xianpeng Lang\inst{2}  \\
    \quad Xiaowei Zhou\inst{1}  
    \quad Sida Peng\inst{1\dagger}  \\
}

\authorrunning{Y. Yan et al.}

\institute{Zhejiang University\inst{1} \quad Li Auto\inst{2}}

\maketitle

\begin{abstract}
  This paper aims to tackle the problem of modeling dynamic urban streets for autonomous driving scenes.
  Recent methods extend NeRF by incorporating tracked vehicle poses to animate vehicles, enabling photo-realistic view synthesis of dynamic urban street scenes. 
  However, significant limitations are their slow training and rendering speed.
  We introduce Street Gaussians, a new explicit scene representation that tackles these limitations. 
  Specifically, the dynamic urban scene is represented as a set of point clouds equipped with semantic logits and 3D Gaussians, each associated with either a foreground vehicle or the background. 
  To model the dynamics of foreground object vehicles, each object point cloud is optimized with optimizable tracked poses, along with a 4D spherical harmonics model for the dynamic appearance. 
  The explicit representation allows easy composition of object vehicles and background, which in turn allows for scene editing operations and rendering at 135 FPS (1066 * 1600 resolution) within half an hour of training. 
  The proposed method is evaluated on multiple challenging benchmarks, including KITTI and Waymo Open datasets. Experiments show that the proposed method consistently outperforms state-of-the-art methods across all datasets. 
  The code will be released to ensure reproducibility.

\end{abstract}

\let\thefootnote\relax\footnotetext{$^{\dagger}$ Corresponding author}

\section{Introduction}
\label{sec:introduction}
Modeling dynamic 3D streets from images has many important applications, such as city simulation, autonomous driving, and gaming.
For instance, the digital twin of city streets can be used as the simulation environment for self-driving vehicles, thereby reducing the training and test costs.
These applications require us to efficiently reconstruct 3D street models from captured data and render high-quality novel views in real-time.

\begin{figure}[t]
    \centering
    \includegraphics[width=1\linewidth]{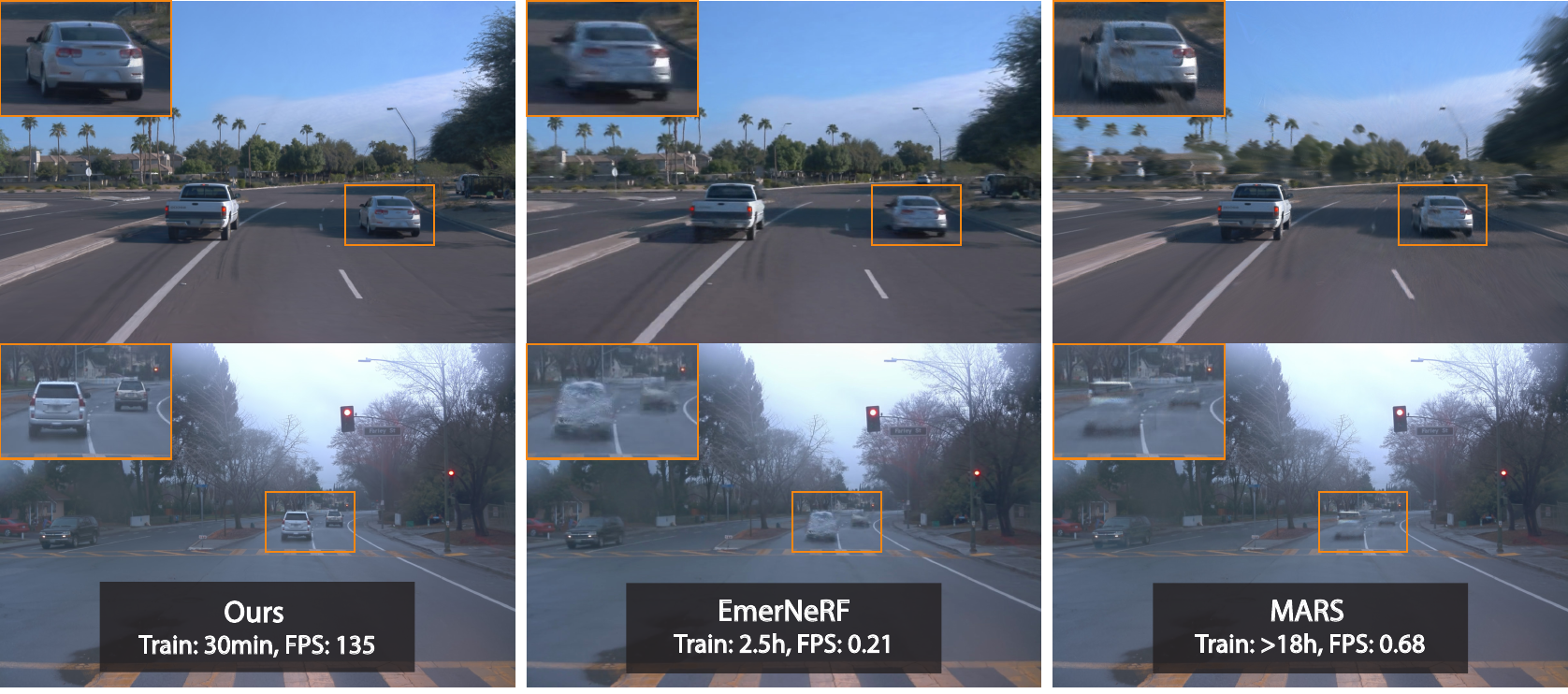}
    \caption{
      \textbf{Rendering results on the Waymo dataset \cite{Sun_2020_CVPR}}. Our method produces high-quality rendering at 135 FPS (1066$\times$1600) within half an hour of training. 
      Current SOTA methods \cite{wu2023mars, yang2023emernerf} suffer from high training and rendering cost. 
    }
    \label{fig:teaser}
\end{figure}

With the development of neural scene representations, there have been some methods \cite{lu2023dnmp,tancik2022block,rematas2022urban,ost2022pointlightfields,zhang2023nerflets} that attempt to reconstruct street scenes with neural radiance fields \cite{mildenhall2020nerf}.
To improve the modeling capability, Block-NeRF \cite{tancik2022block} divides the scene into several blocks and represents each one with a NeRF network.
Although this strategy enables photo-realistic rendering of large-scale street scenes, Block-NeRF suffers from long training time due to the large amount of network parameters.
Moreover, it cannot handle dynamic vehicles on the street, which are crucial aspects in autonomous driving environment simulation.

Recently, some methods \cite{Ost_2021_CVPR, yang2023unisim, wu2023mars, KunduCVPR2022PNF} propose to represent dynamic driving scenes as compositional neural representations that consist of foreground moving cars and static background.
To handle the dynamic car, they leverage tracked vehicle poses to establish the mapping between the observation space and the canonical space, where they use NeRF networks to model the car's geometry and appearance.
Although these methods produce reasonable results, they are still limited to the high training cost and low rendering speed.

In this work, we propose a novel explicit scene representation for reconstructing dynamic 3D street scenes from images.
The basic idea is utilizing point clouds to build dynamic scenes, which significantly increases the training and rendering efficiency. 
Specifically, we decompose urban street scenes into the static background and moving vehicles, which are separately built based on 3D Gaussians \cite{kerbl3Dgaussians}.
To handle the dynamics of foreground vehicles, we model their geometry as a set of points with optimizable tracked vehicle poses, where each point stores learnable 3D Gaussian parameters.
Furthermore, the time-varying appearance is represented by a 4D spherical harmonics model that uses a time series function to predict spherical harmonics coefficients at any time step.
Thanks to the dynamic 3D Gaussians representation, we can faithfully reconstruct the target urban street within half an hour and achieve real-time rendering (135FPS@1066x1600).
Building upon the proposed scene representation, we develop several strategies to further improve the rendering performance, including the tracked pose optimization,point cloud initialization, and sky modeling.

We evaluate the proposed method on Waymo Open \cite{Sun_2020_CVPR} (Waymo) and KITTI \cite{geiger2012we} datasets, 
which present dynamic street scenes with complex vehicle motions and various environment conditions. 
Across all datasets, our approach achieves state-of-the-art performance in terms of rendering quality, 
while being rendered over 100 times faster than previous methods \cite{wu2023mars, Ost_2021_CVPR,yang2023emernerf}. 
Furthermore, detailed ablations and scene editing applications are conducted to 
demonstrate the effectiveness of proposed components and the flexibility of the proposed representation, respectively.

Overall, this work makes the following contributions:
\begin{itemize}
    \item We propose \methodname, a novel scene representation for modeling complex dynamic urban scenes, which efficiently reconstructs and renders high-fidelity urban street scenes in real-time.
    \item We propose several strategies including 4D spherical harmonics appearance model, tracked pose optimization, and point cloud initialization, which largely improve the rendering performance of Street Gaussians.
\end{itemize}

\section{Related work}
\label{sec:related_work}

\PAR{Static scene modeling.}
Neural scene representation proposes to represent 3D scenes with neural networks, which can model complex scenes from images through differentiable rendering.
NeRF \cite{mildenhall2020nerf, barron2021mipnerf, barron2022mipnerf360, barron2023zipnerf, muller2022instant} represents continuous volumetric scenes with MLP networks and achieves impressive rendering results.
Some works have been proposed to extend NeRF to urban scenes \cite{rematas2022urban, tancik2022block, guo2023streetsurf, lu2023dnmp, liu2023neural, ost2022pointlightfields, turki2022mega, irshad2023neo360, cheng2023uc}. 
GridNeRF \cite{xu2023gridguided} proposes muti-resolution feature planes to help NeRF generate photorealistic results on large-scale scenes.
DNMP \cite{lu2023dnmp} models the scene with deformable mesh primitives initialized by voxelizing point clouds. 
NeuRas \cite{liu2023neural} takes scaffold mesh as input and optimizes the neural texture field to perform fast rasterization. 

Point-based rendering works \cite{ruckert2022adop, kopanas2021point, aliev2020neural, dai2020neural, li2022read} define learned neural descriptors
on point clouds and perform differentiable rasterization with a neural renderer. However, they require dense point clouds as input and generate blurry results under regions 
with low point counts. A very recent work 3D Gaussian Splatting (3D GS) \cite{kerbl3Dgaussians} defines a set of anisotropic Gaussians in 3D world and performs adaptive density control
to achieve high-quality rendering results with only sparse point clouds input.  However, 3D GS assumes the scene to be static and can not model dynamic moving objects.

\PAR{Dynamic scene modeling.}
Recent methods build 4D neural scene representation on single-object scenes by encoding time as additional input \cite{song2023nerfplayer, park2021hypernerf, fridovich2023k, li2021neural, attal2023hyperreel, lin2022enerf, peng2023representing, lin2023high}.
Some works learn a scene decomposition of outdoor scenes under the supervision of optical flow \cite{turki2023suds} or vision transformer feature \cite{yang2023emernerf}.
However, their scene representation is not instance aware, limiting the applications for autonomous driving simulation.
Another line of works model the scene as the composition of moving object models and a background model 
\cite{Ost_2021_CVPR, KunduCVPR2022PNF, wu2023mars, yang2023unisim, ziyang2023snerf, tonderski2023neurad} with neural fields, which is most similar to us. 
However, they suffer from high memory cost on large scale scene and can not perform real-time rendering.

Extending point-based rendering to dynamic scene is also investigated recently \cite{xu20234k4d, zhang2022differentiable}. 
Recent approaches extend 3D GS to small-scale dynamic scenes by introducing deformation field \cite{yang2023deformable3dgs, wu20234dgaussians}, 
physical priors \cite{luiten2023dynamic} or 4D parametrization \cite{yang2023gs4d} to 3D Gaussian model.
More recently, some concurrent works \cite{chen2023periodic, zhou2023drivinggaussian} also explore 3D Gaussians in urban street scenes. 
DrivingGaussian \cite{zhou2023drivinggaussian} introduces Incremental 3D Static Gaussians and Composite Dynamic Gaussian
Graphs. PVG \cite{chen2023periodic} utilizes Periodic Vibration 3D Gaussians to model dynamic urban scene.

\PAR{Simulation environments for autonomous driving.}
Existing self-driving simulation engines such as CARLA \cite{dosovitskiy2017carla} or AirSim \cite{shah2018airsim} suffer from costly manual effort to create virtual environments and the lack of realism in the generated data.
In recent years, a lot of effort has been put into building sensor simulations from autonomous driving data captured in real scenes. 
Some works \cite{manivasagam2020lidarsim, yang2020surfelgan, fang2020augmented} concentrate on LiDAR simulation by aggregating LiDAR and reconstructing textured primitives. 
However, they have difficulty handling high-resolution images and usually produce noisy appearance. 
Other works \cite{yang2023reconstructing, chen2021geosim, wang2022cadsim} reconstruct objects from multi-view images and LiDAR input, which can be interacted with other environments. 
However, these methods are restricted to existing images and fail to render novel views.
Some methods utilize neural fields to perform multiply tasks 
including view synthesis \cite{yang2023unisim, Ost_2021_CVPR, huang2023neural},
perception \cite{KunduCVPR2022PNF, fu2022panoptic, zhang2023nerflets}, 
generation \cite{shen2023gina, niemeyer2021giraffe, xu2023discoscene, yang2023urbangiraffe, li2022climatenerf} 
and inverse rendering \cite{wang2022neural, wei2024editable, wang2023neural, pun2023neural} on driving scenes. 
However, they struggle with high training and rendering cost.
In contrast, Our method focuses on performing real-time rendering of dynamic urban scenes, which is crucial for autonomous driving simulation.

\section{Method}
\label{sec:method}

\begin{figure*}[t]
    \centering
    \includegraphics[width=1\linewidth]{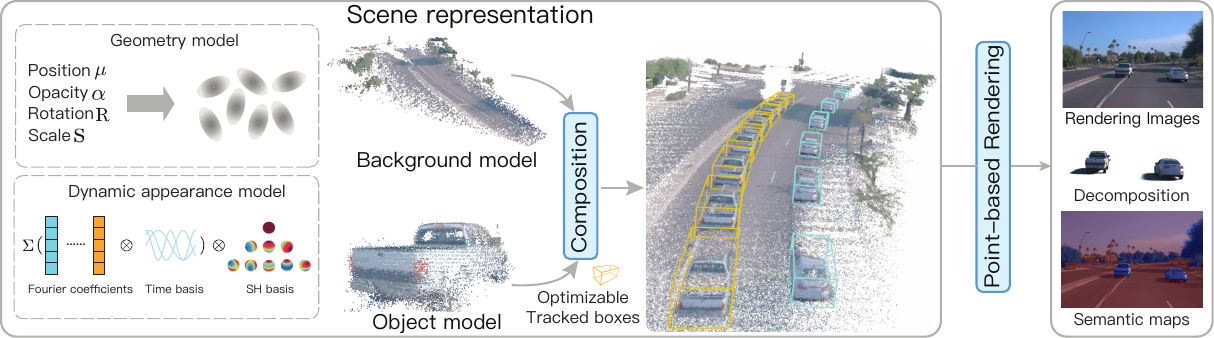}
    \caption{
      \textbf{Overview of \methodname.} 
        The dynamic urban street scene is represented as a set of point-based background and foreground objects with optimizable tracked vehicle poses. 
        Each point is assigned with a 3D Gaussian \cite{kerbl3Dgaussians} including position, opacity, and covariance consisting of rotation and scale to represent the geometry. 
        To represent the appearance, we assign each background point with a spherical harmonics model while the foreground points are associated with a dynamic spherical harmonics model. 
        The explicit point-based representation allows easy composition of separate models,
        which enables real-time rendering of high-quality images and semantic maps (optional if 2D semantic information is provided during training), as well as the decomposition of foreground objects for editing applications. 
    }
    \label{fig:pipeline}
  \end{figure*}

Given a sequence of images captured from a moving vehicle in an urban street scene, our goal is to develop a model capable of generating photorealistic images for view synthesis.
Towards this objective, we propose a novel scene representation, named \methodname, specifically designed for representing dynamic street scenes.
As shown in the Figure \ref{fig:pipeline}, we represent a dynamic urban scene as a set of point clouds, each corresponding to either the static background or a moving vehicle (Section \ref{sec:scene_representation}). 
The explicit point-based representation allows easy composition of separate models, enabling real-time rendering as well as the decomposition of foreground objects for editing applications (Section \ref{sec:rendering}). 
The proposed scene representation can be effectively trained along with tracked vehicle poses from an off-the-shelf tracker, 
enhanced by our pose optimization strategy (Section \ref{sec:training}).  

\subsection{\methodname}
\label{sec:scene_representation}
In this section, we seek to find a dynamic scene representation that can be quickly constructed and rendered in real-time. Previous methods \cite{wu2023mars,KunduCVPR2022PNF} typically face challenges with low training and rendering speed as well as accurate tracked vehicle poses.
To tackle this problem, we propose a novel explicit scene representation, named \methodname, which is built upon 3D Gaussians \cite{kerbl3Dgaussians}.
In \methodname, we represent the static background and each moving vehicle object with a separate neural point cloud. 

In the following, we will first focus on the background model, elaborating on several common attributes that are shared with the object model. Subsequently, we will delve into the dynamic aspects of the object model's design.

\PAR{Background model.}
The background model is represented as a set of points in the world coordinate system.
Each point is assigned with a 3D Gaussian to softly represent the continuous scene geometry and color. 
The Gaussian parameters consist of a covariance matrix $\bm \Sigma_b $ and a position vector $\bm{\mu}_b \in \mathbb{R}^{3}$, which denotes the mean value. 
To avoid invalid value during optimization, each covariance matrix is further reduced to a scaling matrix $\mathbf{S}_b$ and a rotation matrix $\mathbf{R}_b$, where $\mathbf{S}_b$ is characterized by its diagonal elements 
and $\mathbf{R}_b$ is converted into a unit quaternion.
The covariance matrix $\bm{\Sigma}_b $ can be recovered from $\mathbf{S}_b$ and $\mathbf{R}_b$ as:
\begin{equation}
    \label{eq:covariance matrix}
    \bm{\Sigma}_b = \mathbf{R}_b\mathbf{S}_b{\mathbf{S}_b}^T{\mathbf{R}_b}^T.
\end{equation}
Apart from the position and covariance matrix, each Gaussian is also assigned with an opacity value $\alpha_b \in \mathbb{R}$ and a set of
spherical harmonics coefficients $\mathbf{z}_b = (z_{m,l})^{m:-\ell\leq m \leq \ell}_{l: 0\leq \ell \leq \ell_{max}} $ 
to represent scene geometry and appearance.
To obtain the view-dependent color, the spherical harmonics coefficients are further multiplied by the spherical harmonics basis functions projected from the view direction.  
To represent 3D semantic information, each point is added with a semantic logit $\bm{\beta}_b \in \mathbb{R}^M$, where $M$ is the number of semantic classes.

\PAR{Object model.}
Consider a scene containing $N$ moving foreground object vehicles.
Each object is represented with a set of optimizable tracked vehicle poses and a point cloud, where each point is assigned a 3D Gaussian, semantic logits, and a dynamic appearance model.

The Gaussian properties of both the object and the background are similar, sharing the same meaning for opacity $\alpha_o$ and scale matrix $\mathbf{S}_o$. However, their position, rotation, and appearance models differ from those of the background model.
The position $\bm{\mu}_o$ and rotation $\mathbf{R}_o$ are defined in the object local coordinate system.
To transform them into the world coordinate system (the background's coordinate system), we introduce the definition of tracked poses for objects. Specifically, the tracked poses of vehicles are defined as a set of rotation matrices $\{\mathbf{R}_{t}\}_{t=1}^{N_t}$  and translation vectors $\{\mathbf{T}_{t}\}_{t=1}^{N_t}$, where $N_t$ represents the number of frames. The transformation can be defined as:

\begin{equation}
    \label{eq:object gaussian transform}
    \begin{aligned}
    \bm{\mu}_{w} &= \mathbf{R}_t \bm{\mu}_o + \mathbf{T}_t,
    \\
    \mathbf{R}_{w} &= \mathbf{R}_t \mathbf{R}_o,
    \\
    \end{aligned}
\end{equation}
where $\bm{\mu}_w$ and $\mathbf{R}_w$ are the position and rotation of the corresponding object Gaussian in the world coordinate system, respectively. After transformation, the object's covariance matrix $\bm{\Sigma}_w$ can be obtained by Eq. \ref{eq:covariance matrix} with $\mathbf{R}_w$ and $\mathbf{S}_o$.
Note that we also found the tracked vehicle poses from the off-the-shelf tracker to be noisy. To address this issue, we treat the tracked vehicle poses as learnable parameters. We detail it in Section \ref{sec:training}.

\begin{figure}[t]
    \centering
    \includegraphics[width=1\linewidth]{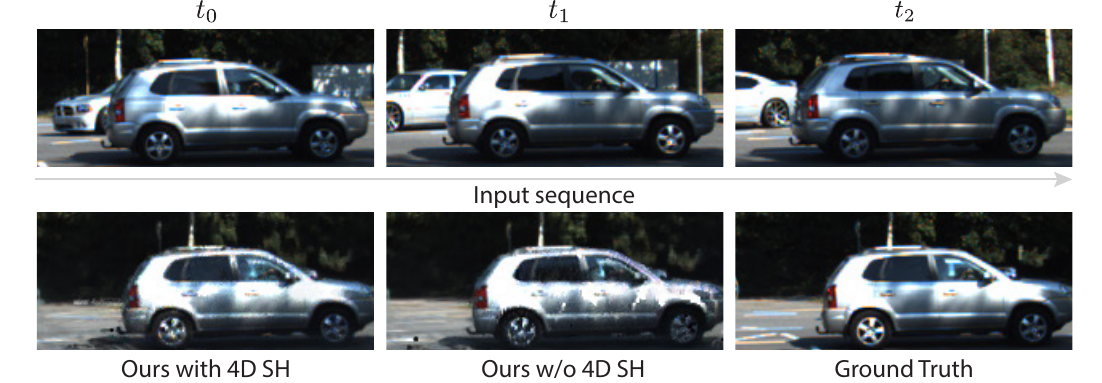}
    \caption{
        \textbf{Effect of 4D SH (spherical harmonics) model.} The first row presents the input sequence, showcasing varying appearances. The second row demonstrates the impact of utilizing the proposed 4D SH model on the rendering results. Significant artifacts can be observed if the 4D SH model is absent. 
    }
    \label{fig:fouriersh}
\end{figure}

Simply representing object appearance with the spherical harmonics coefficients is insufficient for modeling the appearance of moving vehicles, as shown in Figure \ref{fig:fouriersh}, because the appearance of a moving vehicle is influenced by its position in the global scene.
One straightforward solution is to use separate spherical harmonics to represent the object for each timestep. However, this representation will significantly increase the storage cost.
Instead, we introduce the 4D spherical harmonics model by replacing each SH coefficient $z_{m,l}$ with a set of fourier transform coefficients $\bm{f} \in \mathbb{R}^{k}$ where $k$ is the number of fourier coefficient.
Given timestep $t$, $z_{m,l}$ is recovered by performing real-valued Inverse Discrete Fourier Transform:
\begin{equation}
    \label{eq:fourier transform}
    z_{m,l} = \sum_{i=0}^{k-1} \bm{f}_{i} \cos\left(\frac{i\pi}{N_t}t\right).
\end{equation}
With the proposed model, we encode time information into appearance without high storage cost.

The semantic representation of the object model is different from that of background.
The main difference is that the semantic of the object model is a learnable one-dimensional scalar $\beta_o$ which represents the vehicle semantic class from the trakcer 
instead of a $M$-dimensional vector $\bm{\beta}_b$.

\PAR{Initialization.}
The SfM \cite{schonberger2016structure} point cloud used in 3D Gaussian 
is suitable for object centric scene. However, it can not provide good initialization for urban street scenes with many under-observed or textureless regions. 
We instead use aggregated LiDAR point cloud captured by ego vehicle as initialization.
The colors of LiDAR point cloud are obtained by projecting to the corresponding image plane and querying the pixel value.

To initialize the object model, we first collect aggregated points inside the 3D bounding boxes and transform them into the local coordinate system.
For object with less than 2K LiDAR points, we instead randomly sample 8K points inside the 3D bounding box as initialization.
For the background model, we perform voxel downsampling for the remaining point cloud and filter out the points which are invisible to the training cameras. 
We incorporate SfM point cloud to compensate for the limited coverage of LiDAR over large areas. 

\subsection{Rendering of \methodname}
\label{sec:rendering}
To render \methodname, we need to aggregate the contribution of each model to render the final image. Previous methods \cite{Ost_2021_CVPR, KunduCVPR2022PNF, yang2023unisim, wu2023mars}
require compositional rendering with complex raymarching because of neural field representation.
Instead, \methodnameblank can be rendered by contacting all the point clouds and projecting them to 2D image space.
Specifically, given a rendered time step $t$, we first compute spherical harmonics with Eq. \ref{eq:fourier transform} , and transform the object point cloud into the world coordinate system using Eq. \ref{eq:object gaussian transform} according to tracked vehicle pose $(\mathbf{R}_t, \mathbf{T}_t)$. 
Then we concatenate the background point cloud and the transformed object point clouds to form a new point cloud.
To project this point cloud to 2D image space with camera extrinsic $\mathbf{W}$ and intrinsic $\mathbf{K}$, we compute the 2D Gaussian for each point in the point cloud \cite{zwicker2001ewa}:

\begin{equation}
    \begin{aligned}
        \bm{\mu}' &= \mathbf{K} \mathbf{W} \bm{\mu} ,
        \\
        \bm{\Sigma}' &=  \mathbf{J} \mathbf{W} \bm{\Sigma} \mathbf{W}^T  \mathbf{J}^T ,
    \end{aligned}
\end{equation}
where $\mathbf{J}$ is the Jacobian matrix of $\mathbf{K}$ . $\mu'$ and $\Sigma'$ are the position and covariance matrix in 2D image space, respectively. 
Point-based $\alpha$-blending for each pixel is used to compute the color $\mathbf{C}$:
\begin{equation}
    \label{eq:rendering}
    \mathbf{C} = \sum_{i \in N} \mathbf{c}_i \alpha_i \prod_{j=1}^{i-1} (1 - \alpha_j),
\end{equation}
Here $\alpha_i$ is the opacity $\alpha$ multiplied by the probability of the 2D Gaussian and $\mathbf{c}_i$ is the color computed from spherical harmonics $\mathbf{z}$ with the view direction. 
We can also render other signals like depth, opacity and semantic. 
For instance, the semantic map is rendered by changing color $c$ in Eq. \ref{eq:rendering} to semantic logits $\bm{\beta}$.

Since 3D Gaussian is defined in Euclidean space, it is inappropriate for them to model distant regions like sky. 
As a result, we utilize a high resolution cubemap which maps the view direction to sky color $\mathbf{C}_\text{sky}$. 
The explicit cubemap representation
helps us recover details in sky regions without sacrificing inference speed. 
The final rendering color is obtained by blending $\mathbf{C}_\text{sky}$ and the color $\mathbf{C}$ in Eq. \ref{eq:rendering}.
More details can be found in the supplementary.

\subsection{Training}
\label{sec:training}

\PAR{Tracking Pose Optimization.}
Positions and covariance matrices of the object Gaussians during rendering in Section \ref{sec:rendering} are closely correlated with the tracked pose parameters as shown in Eq. \ref{eq:object gaussian transform}. 
However, bounding boxes produced by the tracker model are generally noisy. Directly using them to optimize our scene representation leads to degradation in rendering quality. 
As a result, we treat tracked poses as learnable parameters by adding a learnable transformation to each transformation matrix.  
Specifically, $\mathbf{R}_t $ and $\mathbf{T}_t$ in Eq. \ref{eq:object gaussian transform} are replaced by $\mathbf{R}_t^{'}$ and $\mathbf{T}_t^{'}$ which are defined as:
\begin{equation}
    \label{eq:tracking pose optimization}
    \begin{aligned}
        \mathbf{R}_t' &= \mathbf{R}_t \Delta \mathbf{R}_t,
        \\
        \mathbf{T}_t' &= \mathbf{T}_t + \Delta \mathbf{T}_t,
    \end{aligned}
\end{equation}
where $\Delta \mathbf{R}_t$ and $\Delta \mathbf{T}_t$ are the learnable transformation. 
We represent $\Delta \mathbf{T}_t$ as a 3D vector and
$\Delta \mathbf{R}_i$ as a rotation matrix converted from yaw offset angle $\Delta \theta_t$. 
Gradients of these transformations can be directly obtained without any implicit function or intermediate processes, which do not require any extra computation during back-propagation.

\PAR{Loss function.}
We jointly optimize our scene representation, sky cubemap and tracked poses using the following loss function:

\begin{equation}
    \label{eq:loss function}
    \begin{aligned}
        \mathcal{L} = \mathcal{L}_{\text{color}} + \lambda_1 \mathcal{L}_{\text{depth}} + \lambda_2 \mathcal{L}_{\text{sky}} + 
        \lambda_3 \mathcal{L}_{\text{sem}} + \lambda_4 \mathcal{L}_{\text{reg}}.
    \end{aligned}
\end{equation}

In Eq. \ref{eq:loss function}, $\mathcal{L}_{\text{color}}$ is the reconstruction loss between rendered and observed images following \cite{kerbl3Dgaussians}. 
$\mathcal{L}_{\text{depth}}$ is a L1 loss between rendered depth and the depth generated by projecting sparse LiDAR points onto the camera plane. 
$\mathcal{L}_{\text{sky}}$ is a binary cross entropy loss for sky supervision. 
$\mathcal{L}_{\text{sem}}$ is an optional per-pixel softmax-cross-entropy loss between rendered semantic logits and input 2D semantic segmentation predictions \cite{li2022videoknet} 
and $\mathcal{L}_{\text{reg}}$ is an regularization term used to remove floaters and enhance decomposition effects. 
Please refer to the supplementary material for details of each loss term. 

\section{Implementation details}
\label{sec:implementation_details}

We train \methodnameblank for 30000 iterations with Adam optimizers \cite{kingma2014adam} following the configurations of 3D Gaussians \cite{kerbl3Dgaussians}. The learning rate of translation transformation $\Delta \mathbf{T}_t$ and rotation transformation $\Delta \mathbf{R}_t$ are set to $5e^{-3}$ and $1e^{-3}$, which
decay exponentially to $5e^{-5}$ and $1e^{-5}$ respectively. 
The resolution of sky cubemap is set to 1024 with learning rate decays from  $1e^{-2}$ to $1e^{-4}$ exponentially.
All the experiments are conducted on one single RTX 4090 GPU.

We follow \cite{kerbl3Dgaussians} to apply adaptive control during optimization. 
We fix the scale of background model (20 meters in our experiments) and the scale of each object model is determined by the bounding box dimensions.
In order to prevent object Gaussians from growing to occluded areas, 
for each object model we sample a set of points as a probability distribution function. 
During optimization, Gaussians with sampled points outside the bounding box will be pruned.

\section{Experiments}
\label{sec:experiments}

\subsection{Experimental Setup}

\PAR{Datasets.}
We conduct experiments on Waymo Open Dataset \cite{Sun_2020_CVPR} and KITTI benchmarks \cite{geiger2012we}. 
The frame rates of both datasets are 10 HZ.
On the Waymo Open Dataset, we select 8 recording sequences with large amounts of moving objects, significant ego-car motion and complex lighting conditions. All sequences have a length of around 100 frames. 
We select every 4th image in the sequence as the test frames and use the remaining for training. As we find that our baseline methods \cite{Ost_2021_CVPR,wu2023mars} suffer from high memory cost when training with high-resolution images, we downscale the input images to $1066 \times 1600$. On KITTI \cite{geiger2012we} and Vitural KITTI 2 \cite{cabon2020virtual}, we follow the settings of MARS \cite{wu2023mars} and evaluate our methods with different train/test split settings.
We use the bounding boxes generated by the detector\cite{casa2022} and tracker \cite{wu20213d} on Waymo dataset and use the officially provided object tracklets from KITTI.

\PAR{Baseline methods.}
We compare our methods with four recent methods. 
(1) NSG \cite{Ost_2021_CVPR} represents background as multi-plane images and use per-object learned latent codes with a shared decoder to model moving objects. 
(2) MARS \cite{wu2023mars} builds the neural scene graph based on Nerfstudio \cite{nerfstudio}.
(3) 3D Gaussians \cite{kerbl3Dgaussians} models the scene with a set of anisotropy gaussians.
(4) EmerNeRF \cite{yang2023emernerf} stratifies scenes into static and dynamic fields, each modeled with a hash grid \cite{muller2022instant}.
Both NSG and MARS are trained and evaluated using ground truth object tracklets. Details of baseline implementations can be found in the supplementary.

\begin{figure}[!htbp]
    \centering
    \includegraphics[width=1\linewidth]{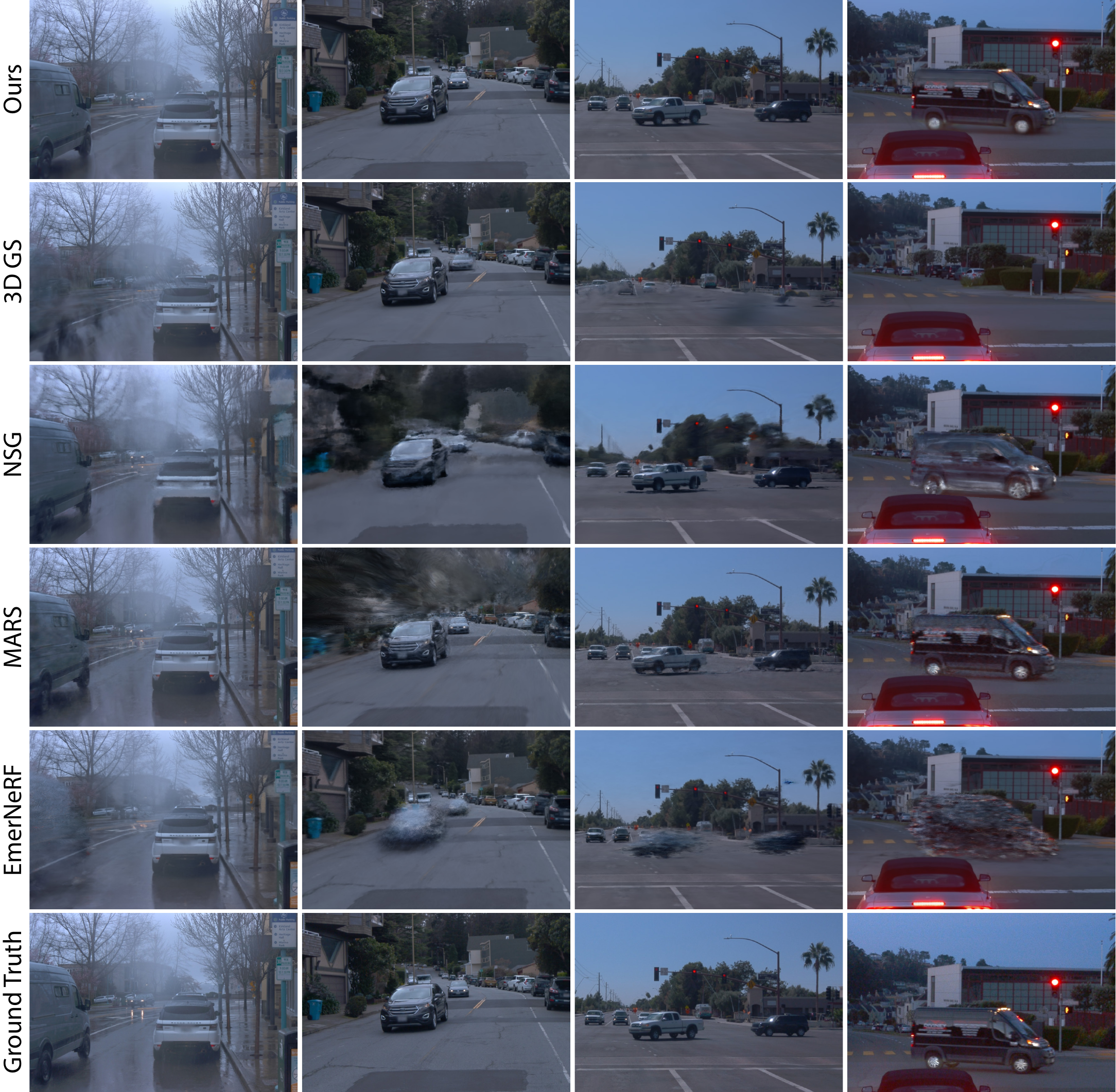}
    \caption{
        \textbf{Qualitative comparisons results on the Waymo \cite{Sun_2020_CVPR} dataset.}  NSG \cite{Ost_2021_CVPR} and MARS \cite{wu2023mars} often produce blurry and distorted results. 3D GS \cite{kerbl3Dgaussians} and EmerNeRF 
        \cite{yang2023emernerf} generates ghosting artifacts in regions with moving objects. In contrast, our approach significantly outperforms other methods with high fidelity and sharp details. 
    }   
    \label{fig:nvs_waymo}
\end{figure}

\begin{table}
    \centering
    \caption{\textbf{
    Quantitative results on the Waymo \cite{Sun_2020_CVPR} dataset.} The rendering image resolution is $1066 \times 1600$. ``PSNR*'' denotes
        the PSNR of moving objects.}
    \scalebox{1.0}{
        \begin{tabular}{lccccc}
            \toprule
            
            & 3D GS \cite{kerbl3Dgaussians} & NSG \cite{Ost_2021_CVPR} & MARS \cite{wu2023mars} & EmerNeRF \cite{yang2023emernerf} & Ours \\
            \midrule
            PSNR$\uparrow$ & 29.64 & 28.31 & 29.75 & 30.87& \textbf{34.61}    \\ 
            SSIM$\uparrow$ & 0.918 & 0.862 & 0.886 & 0.905 & \textbf{0.938}     \\
            LPIPS$\downarrow$ & 0.117 & 0.346 & 0.264 & 0.133 & \textbf{0.079}    \\ 
            \midrule
            PSNR*$\uparrow$ & 21.25 & 24.32 & 26.54 & 21.67 & \textbf{30.23}    \\
            \midrule
            FPS$\uparrow$ & \textbf{205} & 0.47 & 0.68 & 0.21 & 135 \\
            \bottomrule
            \end{tabular}
            
    }


    \label{tab:nvs_waymo}
    \end{table}

\begin{table*}[t!]
    \centering
    \caption{\textbf{Quantitative results on KITTI \cite{geiger2012we} and VKITTI2 \cite{cabon2020virtual} datasets.} We strictly follow the experimental setting of MARS \cite{wu2023mars} and borrow results of MARS \cite{wu2023mars} and NSG \cite{Ost_2021_CVPR} from it. The rendering image resolution is 375 $\times$ 1242. 
    }
    \scalebox{0.85}{    \begin{tabular}{lcccccccccccc}
        \toprule
        && \multicolumn{3}{c}{KITTI - 75\%} && \multicolumn{3}{c}{KITTI - 50\%} && \multicolumn{3}{c}{KITTI - 25\%} \\
        \cmidrule{3-5} \cmidrule{7-9} \cmidrule{11-13}
        && PSNR$\uparrow$ & SSIM$\uparrow$& LPIPS$\downarrow$ && PSNR$\uparrow$ & SSIM$\uparrow$& LPIPS$\downarrow$ && PSNR$\uparrow$ & SSIM$\uparrow$& LPIPS$\downarrow$  \\
        \midrule
        3D GS \cite{kerbl3Dgaussians} && 19.19 & 0.737 & 0.172 && 19.23 & 0.739 & 0.174 && 19.06 & 0.730 & 0.180 \\
        NSG* \cite{Ost_2021_CVPR} && 21.53 & 0.673 & 0.254 && 21.26 & 0.659 & 0.266 && 20.00 & 0.632 & 0.281 \\
        MARS* \cite{wu2023mars} && 24.23 & \textbf{0.845} & 0.160 && 24.00 & 0.801 & 0.164 && 23.23 & 0.756 & 0.177 \\         
        Ours && \textbf{25.79} & 0.844 &  \textbf{0.081} &&  \textbf{25.52} & \textbf{0.841} & \textbf{0.084}  &&  \textbf{24.53} & \textbf{0.824} & \textbf{0.090} \\
        \midrule
        && \multicolumn{3}{c}{VKITTI2 - 75\%} && \multicolumn{3}{c}{VKITTI2 - 50\%} && \multicolumn{3}{c}{VKITTI2 - 25\%} \\    \cmidrule{3-5} \cmidrule{7-9} \cmidrule{11-13}
        && PSNR$\uparrow$ & SSIM$\uparrow$& LPIPS$\downarrow$ && PSNR$\uparrow$ & SSIM$\uparrow$& LPIPS$\downarrow$ && PSNR$\uparrow$ & SSIM$\uparrow$& LPIPS$\downarrow$  \\
        \midrule
        3D GS \cite{kerbl3Dgaussians} && 21.12 & 0.877 & 0.097 && 21.11 & 0.874 & 0.097 && 20.84 & 0.863 & 0.098 \\
        NSG* \cite{Ost_2021_CVPR} && 23.41 & 0.689 & 0.317 && 23.23 & 0.679 & 0.325 && 21.29 & 0.666 & 0.317 \\
        MARS* \cite{wu2023mars} && 29.79 & 0.917 & 0.088 && 29.63 & 0.916 & 0.087 && 27.01 & 0.887 & 0.104 \\
        Ours && \textbf{30.10} & \textbf{0.935} &  \textbf{0.025} &&  \textbf{29.91} & \textbf{0.932} &  \textbf{0.026} && \textbf{28.52} & \textbf{0.917} & \textbf{0.034} \\
        \bottomrule
        \end{tabular}}

    \label{tab:nvs_kitti_vkitti}
\end{table*}

\subsection{Comparisons with the State-of-the-art}
Tables \ref{tab:nvs_waymo}, \ref{tab:nvs_kitti_vkitti} present the comparison results of our method with baseline methods \cite{kerbl3Dgaussians, Ost_2021_CVPR, wu2023mars, yang2023emernerf} in terms of rendering quality and rendering speed. We adopt PSNR, SSIM and LPIPS \cite{Zhang_2018_CVPR} as metrics to evaluate rendering quality. 
To better evaluate the rendering quality of moving objects, we project 3D bounding boxes to 2D image plane and calculate the loss only on pixels inside the projected box, 
which is denoted as PSNR* in our experiments. 
For all the metrics, our model achieves the best performance among all the methods with a 12.1\% increase in PSNR and a 13.9\% increase in PSNR*.
Moreover, our method renders two magnitudes faster than NeRF-based methods \cite{Ost_2021_CVPR, wu2023mars, yang2023emernerf}. 
Although 3D GS is faster than our method, it can only support static scenes and the rendering result of moving objects degrades significantly. 

Figure \ref{fig:nvs_waymo} shows the qualitative results of our method and baselines on the Waymo dataset.
3D GS fails to model dynamic objects and EmerNeRF can not generates reasonable results in dynamic regions of novel timestep. 
Although given ground truth tracking poses, NSG and MARS still suffer from blurry and distorted results due to the lack of capacity of
their model when the scene is complex. 
In contrast, our method can generate high-quality novel views with high fidelity and details.

\begin{table}
    \centering
    \caption{\textbf{Ablation studies on the Waymo \cite{Sun_2020_CVPR} dataset.} Metrics are averaged over all the sequences on the Waymo dataset.  ``PSNR*'' denotes the PSNR of moving objects. ``opt.'' denotes optimization. Please refer to Section \ref{sec:ablation} for details. 
    }
    \scalebox{1.0}{
        \begin{tabular}{lcccc}
            \toprule
            
            & PSNR$\uparrow$  & PSNR*$\uparrow$  & SSIM$\uparrow$ & LPIPS$\downarrow$ \\
            \midrule
            Ours w/o LiDAR & 34.02 & 29.53 & 0.934 & 0.087 \\
            Ours w/o 4DSH & 34.36 & 29.27  & 0.937 & 0.081 \\
            Ours w/o pose opt. & 34.18 & 28.24  & 0.935 & 0.081 \\
            Ours w/ GT pose & \textbf{34.61} & 29.84 & 0.937 & 0.080 \\
            Complete model & \textbf{34.61} & \textbf{30.23} & \textbf{0.938} & \textbf{0.079}  \\
            \bottomrule
            \end{tabular}       
    }

    \label{tab:ablation}

    \end{table}
    
\begin{figure}
    \centering
    \includegraphics[width=1\linewidth]{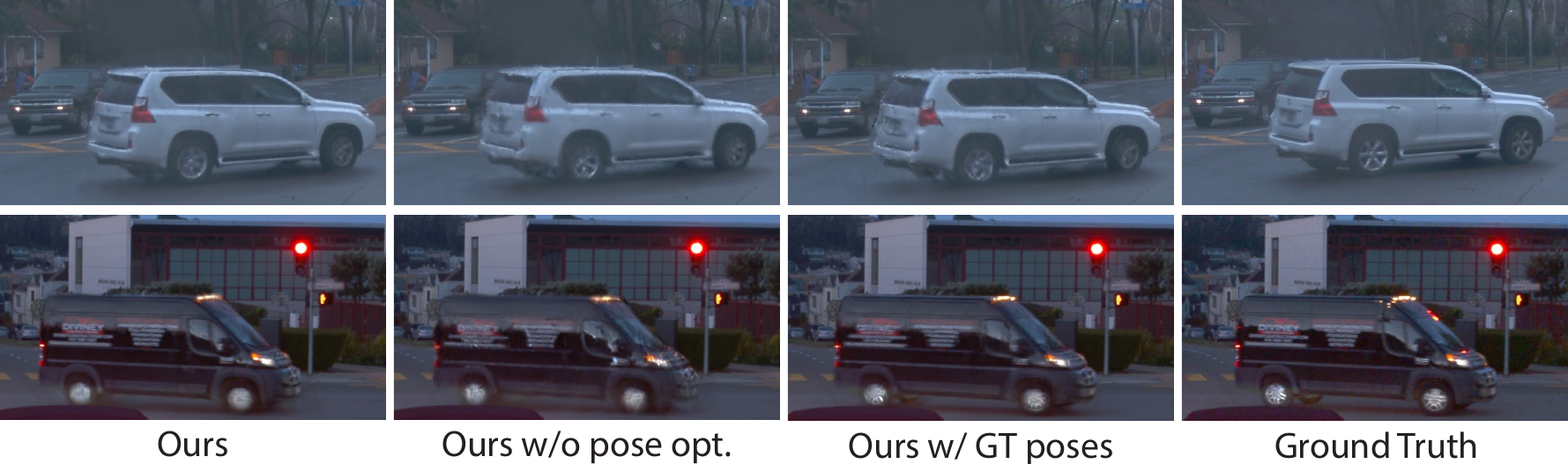}
    \caption{
        \textbf{Ablation study on tracking pose optimization.} The results indicate that optimizing tracked poses improves the quality. ``opt.'' denotes optimization.
    }
    \label{fig:ablation_pose_opt}
\end{figure}

\begin{figure}
    \centering
    \includegraphics[width=1\linewidth]{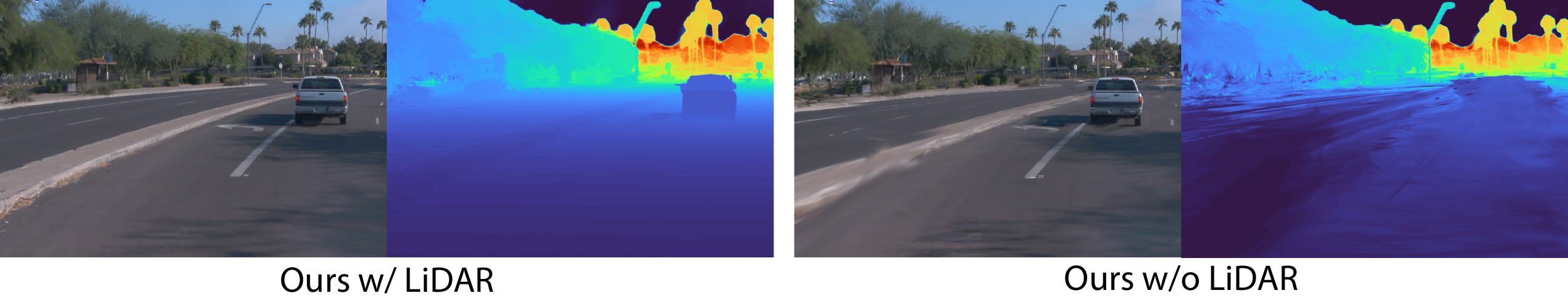}
    \caption{
        \textbf{Ablation study on LiDAR point cloud.} We show the rendered image and depth of our method with and without LiDAR as input. 
    }
    \label{fig:ablation_lidar}
\end{figure}

\subsection{Ablations and Analysis}
\label{sec:ablation}
We validate our algorithm's design choices on all selected sequences from the Waymo dataset. Table \ref{tab:ablation} presents the quantitative results.

\PAR{Importance of optimizing tracked poses.}
Experimental results in Table \ref{tab:ablation} show that our complete model outperforms the model trained without tracking pose optimization by a large margin, 
which indicates the effectiveness of our pose optimization strategy. 
It is interesting to notice that the result of our method is even better than the model trained with ground truth poses, 
a plausible explanation is that there still exists noise in ground truth annotations.

Visual results of the influence of tracked pose optimization is shown in Figure \ref{fig:ablation_pose_opt}. 
Treating tracked poses as learnable parameters help the object model synthesize more texture details like the rear of the white vehicle or the logo of the black vehicle and reduce rendering artifacts.

\PAR{Effectiveness of 4D spherical harmonics.}
Results in Table \ref{tab:ablation} indicate that our 4D spherical harmonics appearance model can refine the rendering quality.
This situation becomes particularly evident when the object interacts with environmental lighting as shown in Figure \ref{fig:fouriersh}.
Our model can generate smooth shadows on the car while the rendering results without 4D spherical harmonics are much noisier.  

\PAR{Influence of incorporating LiDAR points.} 
We evaluate the influence of LiDAR point cloud by comparing our method to a variant 
with SfM initialization for background and random initialization for moving object as described in Section \ref{sec:scene_representation}. 
We also disable the LiDAR depth loss in  Eq. \ref{eq:loss function}.
Table \ref{tab:ablation} shows that incorporating LiDAR point cloud enhances the results of both background and moving objects.  
Figure \ref{fig:ablation_lidar} indicates that using LiDAR points helps our model recover more accurate scene geometry and reduce blurry artifacts. 
It is worth noticing that our method still significantly outperforms baseline methods even without LiDAR input as shown in Tables \ref{tab:ablation}, \ref{fig:nvs_waymo}, 
which proves the efficiency of \methodnameblank under different settings.  

\begin{figure}[t]
    \centering
    \includegraphics[width=1\linewidth]{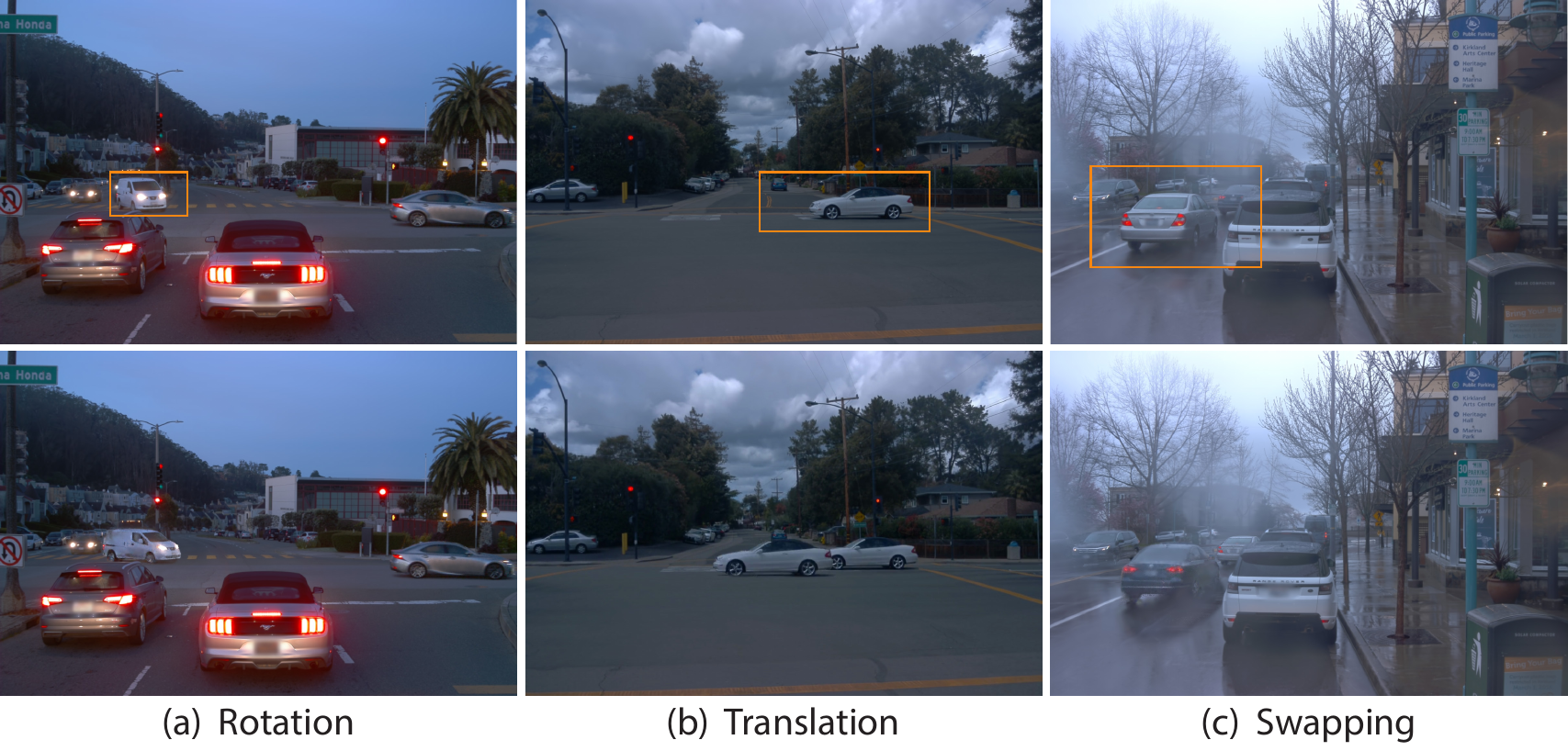}
    \caption{
        \textbf{Editing operations on the Waymo  \cite{Sun_2020_CVPR} dataset.} 
        Images in the first and second rows represent the results before and after editing.
        Our method supports various editing operations, including rotation, translation and swapping.
    }
    \label{fig:waymo_editing}
\end{figure}

\begin{figure}[t]
    \centering
    \includegraphics[width=1\linewidth]{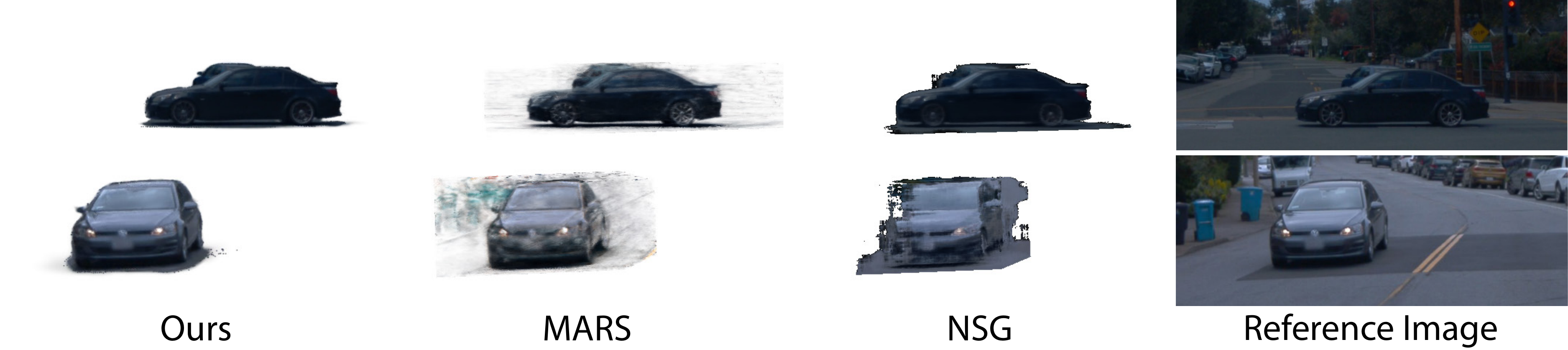}
    \caption{
        \textbf{Decomposition results on the Waymo \cite{Sun_2020_CVPR} dataset.} 
        NSG \cite{Ost_2021_CVPR} cannot decompose clean foreground objects while MARS \cite{wu2023mars} generates floaters in background regions. 
        Instead, our method successfully decomposes the foreground objects and produces high fidelity rendering results.
    }
    \label{fig:waymo_decompostion}
\end{figure}

\begin{table*}[t!]
    \centering
    \caption{\textbf{Quantitative segmentation results on the KITTI \cite{geiger2012we} dataset.} 
    ``VKN ground-truth'' and ``VKN rendered'' denote semantic prediction results of Video K-Net with ground-truth images and our rendered images, respectively.
    }
    \scalebox{0.9}{
        \begin{tabular}{lccccccc}
            \toprule
            
            Method & VKN ground-truth & VKN rendered & Ours \\
            \midrule
            mIoU $\uparrow$ &  57.94 & 53.81 & \textbf{58.81} \\
            \bottomrule
            \end{tabular}       
    }

    \label{tab:semantic_kitti}
\end{table*}

\begin{figure}
    \centering
    \includegraphics[width=1\linewidth]{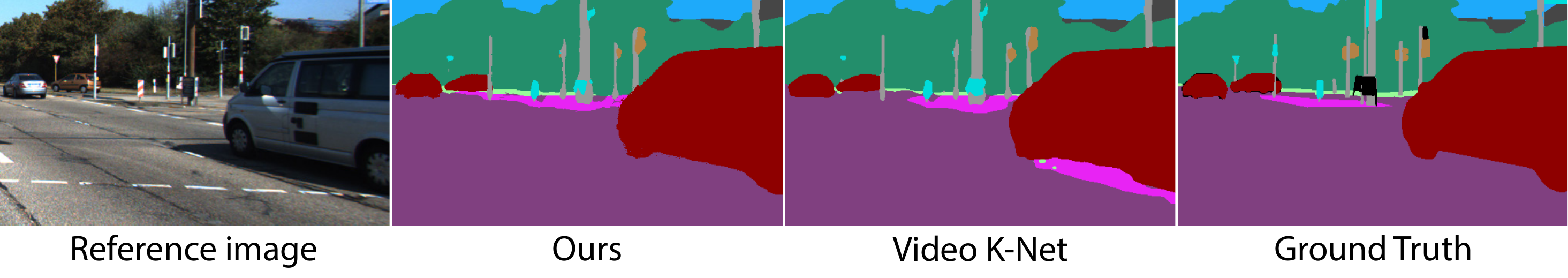}
    \caption{
        \textbf{Visual semantic segmentation results on the KITTI \cite{geiger2012we} dataset.} 
        It can be observed that our method achieves better performance, particularly in ambiguous areas such as shadows, due to our ability to fuse semantic information in 3D.
    }
    \label{fig:semantic_kitti}
\end{figure}

\subsection{Applications}
\methodnameblank can be applied to multiple tasks in computer vision including object decomposition, semantic segmentation and scene editing.

\PAR{Scene editing.}
Our instance-aware scene representation enables various types of scene editing operations.
We can rotate the heading of the vehicle (Figure \ref{fig:waymo_editing} (a)), translate the vehicle (Figure \ref{fig:waymo_editing} (b)) 
and swap one vehicle in the scene with another one (Figure \ref{fig:waymo_editing} (c)).

\PAR{Object Decomposition.}
We compare the decomposition results of our method with NSG \cite{Ost_2021_CVPR} and MARS \cite{wu2023mars} under the Waymo dataset. 
As shown in Figure \ref{fig:waymo_decompostion},
NSG fails to disentangle foreground objects from the background and the result of MARS is blurry due to the model capacity
and lack of regularization. In contrast, our method can produce high-fidelity and clean decomposed results.

\PAR{Semantic Segmentation.}
We compare the quality of our rendered semantic map with the semantic prediction from Video-K-Net \cite{li2022videoknet} on KITTI dataset. 
Our semantic segmentation model is trained with results from Video K-Net. Qualitative and quantitative results
are shown in Figure \ref{fig:semantic_kitti} and Table \ref{tab:semantic_kitti}. Our semantic maps achieve better performance thanks to our representation.

\section{Conclusion}
\label{sec:conclusion}
This paper introduced \methodname, an explicit scene representation for modeling dynamic urban street scenes.
The proposed representation separately models the background and foreground vehicles as a set of neural point clouds.
This explicit representation allows easy compositing of object vehicles and background, enabling scene editing and real-time rendering within half an hour of training.
Furthermore, we demonstrate that the proposed scene representation can achieve comparable performance to that achieved using precise ground-truth poses, using only poses from an off-the-shelf tracker. 
Detailed ablation and comparison experiments are conducted on several datasets, demonstrating the effectiveness of the proposed method.

\PAR{Acknowledgement.} The authors would like to acknowledge the support from NSFC (No. 623B2091), Li Auto and Information Technology Center and State Key Lab of CAD\&CG, Zhejiang University.

%
%
\bibliographystyle{splncs04}
\bibliography{main}

\clearpage

\appendix

\section{More implementation details}

\subsection{Street Gaussians implementations.}

\PAR{Point cloud initialization.}
We obtain SfM point cloud of background model by treating camera poses as known parameters and perform point triangulation. 
As moving objects violate the assumption of multi-view consistency, we ignore these parts by using the mask as shown in Figure \ref{fig:psnr_obj}
during feature extraction. We can directly concatenate SfM and LiDAR point cloud as they are both defined in the world coordinate system.

\PAR{Object semantic.}
To merge the one-dimensional scalar $\beta_o$ with the M-dimensional vector $\beta_b$ of background, we convert 
$\beta_o$ to a M-dimensional one-hot vector for the vehicle label during rendering.

\PAR{Sky cubemap.}
The sky cubemap takes viewing direction $\mathbf{d}$ as input and output sky color $\mathbf{C}_\text{sky}$. Let the rendered color and opacity
of Gaussians as $\mathbf{C}_g$ and $\mathbf{O}_g$, the final rendering color $\mathbf{C}$ can be written as:
\begin{equation}
    \label{eq:blender color}
    \begin{aligned}
        \mathbf{C} = \mathbf{C}_g + (1 - \mathbf{O}_g) * \mathbf{C}_\text{sky}.
    \end{aligned}
\end{equation}

\PAR{Loss functions.}
As we discussed in the main paper, our total loss function is:
\begin{equation}
    \label{eq:loss function}
    \begin{aligned}
        \mathcal{L} = \mathcal{L}_{\text{color}} + \lambda_1 \mathcal{L}_{\text{depth}} + \lambda_2 \mathcal{L}_{\text{sky}} + 
        \lambda_3 \mathcal{L}_{\text{sem}} + \lambda_4 \mathcal{L}_{\text{reg}}.
    \end{aligned}
\end{equation}

\begin{figure}[t]
    \centering
    \includegraphics[width=0.8\linewidth]{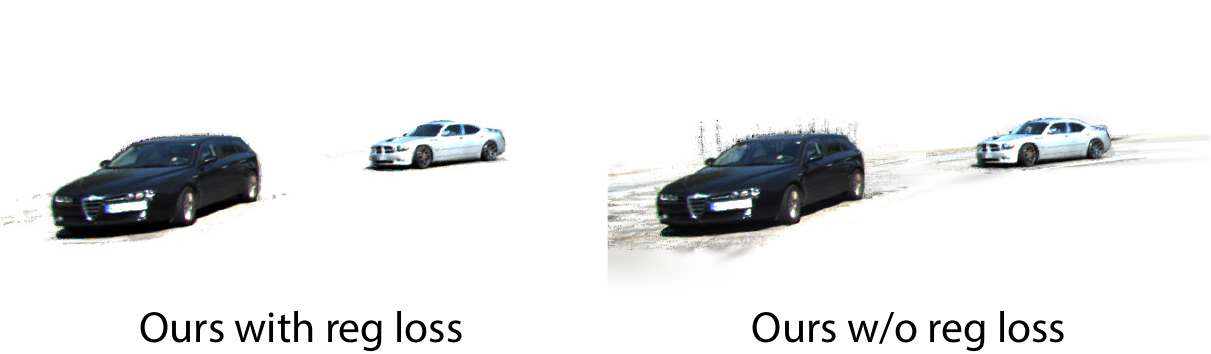}
    \caption{
        \textbf{Effect of regularization loss on decomposition results}. ``reg loss'' denotes regularization loss. 
        Adding this term can significantly remove ghosty artifacts around the vehicle.
    } 
\label{fig:reg_loss}
\end{figure}

\begin{enumerate}
    \item $\mathcal{L}_{\text{color}}.$ We apply the $\mathcal{L}_1$ and D-SSIM loss between rendered and observed images:
    \begin{equation}
        \label{eq:color loss}
        \begin{aligned}
            \mathcal{L}_{\text{color}} = (1 - \lambda_{\text{SSIM}}) \mathcal{L}_1 + \lambda_{\text{SSIM}} \mathcal{L}_{\text{D-SSIM}}.
        \end{aligned}
    \end{equation}
    We set $\lambda_{\text{SSIM}}$ to 0.2 following \cite{kerbl3Dgaussians}.

    \item $\mathcal{L}_{\text{depth}}.$ We apply the $\mathcal{L}_1$ loss between rendered depth $\mathbf{D}$ and the LiDAR measurement's depth $\mathbf{D}^{\text{lidar}}$:
    \begin{equation}
        \label{eq:depth loss}
        \begin{aligned}
            \mathcal{L}_{\text{depth}} = \sum || \mathbf{D} - \mathbf{D}^{\text{lidar}} ||_1
        \end{aligned}
    \end{equation}
    We optimize 95\% of the pixels with smallest depth error to prevent noisy LiDAR observations from affecting the optimization \cite{yang2023unisim}. $\lambda_1$ is set to 0.01.
    
    \item $\mathcal{L}_{\text{sky}}.$ We apply the binary cross entropy loss between rendered opacity $\mathbf{O}_g$ and predicted sky mask $\mathbf{M}_{\text{sky}}$:
    \begin{equation}
        \label{eq:sky loss}
        \begin{aligned}
            \mathcal{L}_{\text{sky}} = -\sum ((1 - \mathbf{M}_{\text{sky}}) \text{log} \mathbf{O}_g + \mathbf{M}_{\text{sky}} \text{log}(1 - \mathbf{O}_g))
        \end{aligned}
    \end{equation}
    $\mathbf{M}_{\text{sky}}$ is generated by Grounded SAM \cite{ren2024grounded}. To be specific, we first get 2D boxes by entering text "sky" to Grounding Dino \cite{liu2023grounding}.
    Then we input the boxes as prompt to SAM \cite{kirillov2023segany} and obtain the predicted sky mask. $\lambda_2$ is set to 0.05.

    \item $\mathcal{L}_{\text{sem}}.$ We apply the per-pixel softmax-cross-entropy loss between rendered semantic logits and predicted 2D semantic segmentation \cite{li2022videoknet}.
    In order to prevent noisy input semantic labels from influencing the scene geometry \cite{Siddiqui_2023_CVPR}, we only perform backpropagation to semantic logits $\beta$ for $\mathcal{L}_{\text{sem}}$.
    $\lambda_3$ is set to 0.1.

    \item $\mathcal{L}_{\text{reg}}.$ The regularization term in our loss function is defined as 
    an entropy loss on the accumulated alpha values of decomposed foreground objects $\mathbf{O}_{\text{obj}}$: 
    \begin{equation}
        \label{eq:reg loss}
        \begin{aligned}
            \mathcal{L}_{\text{reg}} = -\sum (\mathbf{O}_{\text{obj}} \text{log} \mathbf{O}_{\text{obj}} + (1 - \mathbf{O}_{\text{obj}}) \text{log} (1 - \mathbf{O}_{\text{obj}}))
        \end{aligned}
    \end{equation}
    We add this loss after the adaptive control process to help our model better distinguishes foreground and background. 
    Figure \ref{fig:reg_loss} shows the qualitative results, which demonstrates the effect of this regularization term. $\lambda_4$ is set to 0.1.
\end{enumerate}

\PAR{Hyperparameters.}
In practice, we set the number of fourier coefficients $k$ as 5 to maintain a balance between performance and storage cost.
Due to the relatively less intense view-dependent effect on urban scene compared to dataset in 3D Gaussian \cite{kerbl3Dgaussians}, 
we reduce the SH degree to 1 to prevent overfitting. 
We set the voxel size to 0.15m when performing downsampling for LiDAR point cloud.

\subsection{Baselines implementations.}
We give detailed descriptions of our baseline implementations.
\begin{enumerate}
    \item Neural Scene Graph \cite{Ost_2021_CVPR}. We use the official implementation and try a variant where each moving object is modeled
        by a separate NeRF network instead of a shared decoder. The best result is reported for each scene.
    \item Mars \cite{wu2023mars}. We use the official implementation and try a variant where each moving object is modeled
        by a separate Nerfacto \cite{nerfstudio} model. We choose the appearance embedding of nearest training frame as input to each test frame.
        The best result is reported for each scene.
    \item 3D Gaussians \cite{kerbl3Dgaussians}. We initialize the point cloud by running Colmap \cite{schonberger2016structure} with known camera parameters from the dataset. 
        We find that the number of SfM point cloud generated by Colmap is usually less than 1K when the ego-vehicle has little motion. 
        We use LiDAR points to initialize 3D Gaussians for these cases to get reasonable results.
    \item EmerNeRF \cite{yang2023emernerf}. We run the official code under the setting incorporating dynamic encoder, flow encoder and feature lifting.
        We use the same sky mask obtained from Grounded SAM \cite{ren2024grounded} for fair comparison.
\end{enumerate}

\subsection{Evaluations}
\begin{figure}[t]
    \centering
    \includegraphics[width=0.8\linewidth]{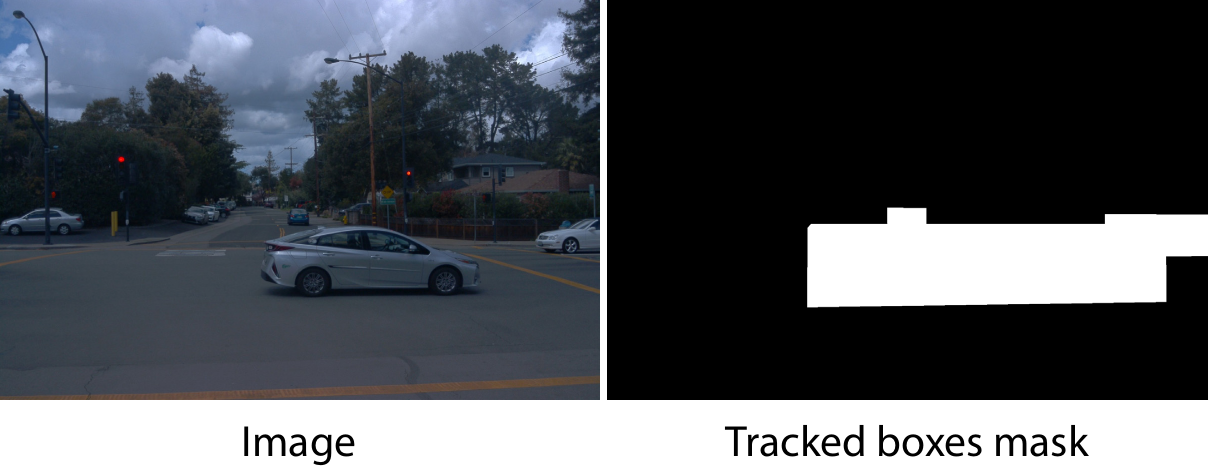}
    \caption{
        \textbf{Illustration of PSNR*}. We project the 3D tracked boxes to 2D image plane
        and obtain the mask above. We calculate the MSE (Mean Squared Error) for the pixels within the mask to get the value of PSNR*.
    } 
\label{fig:psnr_obj}
\end{figure}

Figure \ref{fig:psnr_obj} visually illustrates the calculation method of the PSNR* metric in our experiments. We expand each 
bounding box by 1.5 times in both length and width dimensions to ensure it fully covers the object.
For fair comparison, both our method and the baselines are evaluated using the mask obtained from object tracklets provided by the dataset.

\section{Additional experiments}

\PAR{Analysis of optimizing tracked poses.}
As discussed in our main paper, we observe that our explicit representation facilitates the optimization of tracked vehicle poses with ease. 
Herein, we extend our study to explore the impact of an implicit representation \cite{wu2023mars} on optimizing tracked vehicle poses. 
The experimental results, as presented in Table \ref{tab:pose opt}, indicate that while the inclusion of our pose optimization strategy with implicit representation improves outcomes, 
there remains a noticeable gap compared to experiments using ground truth tracked poses. 
However, the proposed method, employing tracked poses from an off-the-shelf tracker, achieves results comparable to those using GT poses. 
This success can be attributed to the more efficient propagation of gradients through explicit representations in relation to tracked poses.

\begin{table}
    \centering
    \caption{\textbf{More ablation studies on tracking pose optimization.} We report the results of PSNR* on two scenes from Waymo dataset. ``opt.'' denotes optimization.
    }
    \scalebox{0.9}{
        \begin{tabular}{lccc}
            \toprule
            & \multicolumn{3}{c}{Sequence A} \\ \cmidrule{2-4} 
            & w/o pose opt.  & with pose opt. & with GT poses  \\
            \midrule
            MARS \cite{wu2023mars} & 25.78 & 27.68 & 29.74 \\
            Ours & 29.08 & 31.35 & 30.84 \\
            \midrule
            & \multicolumn{3}{c}{Sequence B} \\ \cmidrule{2-4} 
            & w/o pose opt.  & with pose opt. & with GT poses  \\
            \midrule
            MARS \cite{wu2023mars} & 24.38 & 25.83 & 26.94 \\
            Ours & 25.86 & 27.98 & 28.02 \\
            \bottomrule
            \end{tabular}       
    }

    \label{tab:pose opt}
    \vspace{-3mm}

\end{table}

\begin{figure}[t]
    \centering
    \includegraphics[width=0.9\linewidth]{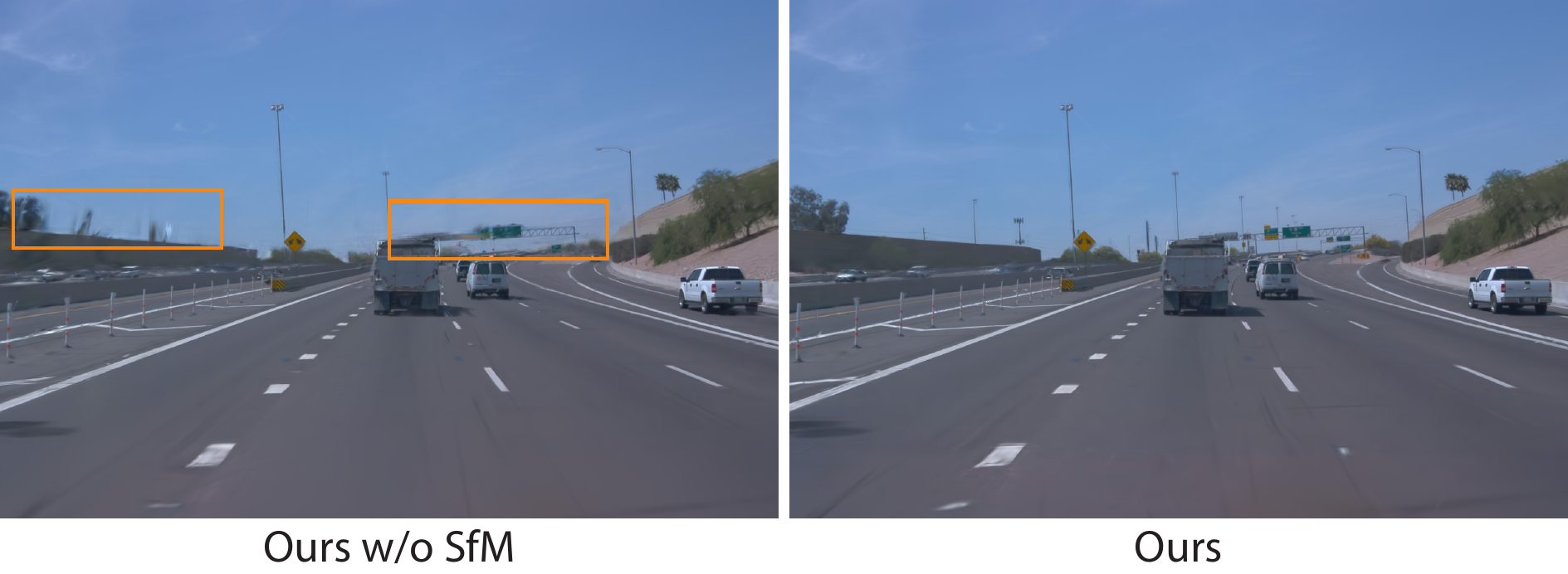}
    \caption{
        \textbf{Effect of incorporating SfM points on novel view synthesis results.}
    } 
\label{fig:ablation_colmap}
\end{figure}

\PAR{Analysis of point cloud initialization.}
We have demonstrated the importance of including LiDAR point cloud during initialization. However, the LiDAR points can not cover the entire scene, especially for far away 
regions. As a result, the SfM point cloud is also crucial for the reconstruction of dynamic urban scene. 
We perform an ablation study by using only LiDAR point cloud to initialize the background model. 
As shown in Figure \ref{fig:ablation_colmap}, although LiDAR points can help recover texture details in near regions like the road, 
it cannot restore some areas not covered by LiDAR points, such as trees on the other side of the road or distant road signs.
The rendering results without SfM points is even worse than the one without LiDAR points as illustrated in Table \ref{tab:colmap_sky}.    
Our approach combines them as input to leverage their respective strengths.

\begin{figure}[t]
    \centering
    \includegraphics[width=0.9\linewidth]{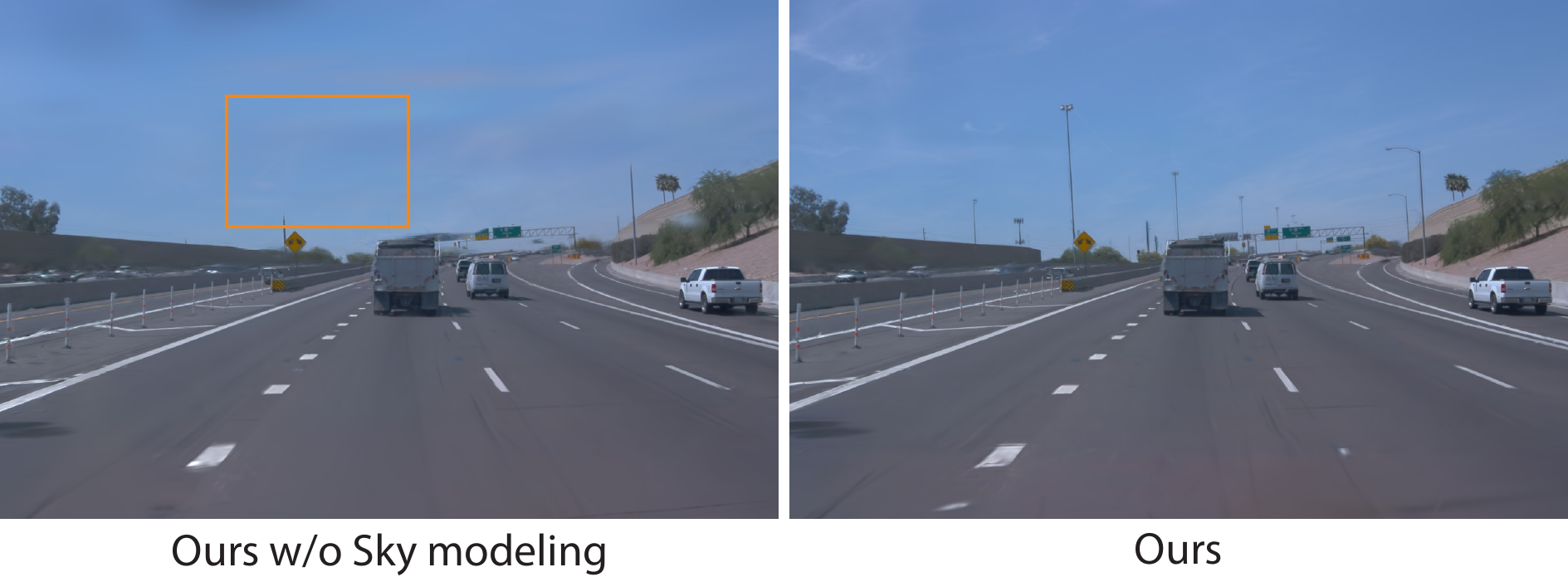}
    \caption{
        \textbf{Effect of modeling sky with cubemap on novel view synthesis results.}
    } 
\label{fig:ablation_sky}
\end{figure}

\PAR{Analysis of sky modeling.}
As shown in Figure \ref{fig:ablation_sky} and Table \ref{tab:colmap_sky}, using a separate cube map to model the sky can help better recover detail areas, 
while avoiding some foreground objects being obscured by the gaussians representing sky regions.

\begin{table}
    \centering
    \caption{
        \textbf{More ablation studies on point cloud initialization and sky modeling.} We show the quantitative results on two scenes from Waymo dataset 
        with large scale background and many thin structures.
    }
    \scalebox{1.0}{
        \begin{tabular}{lcccc}
            \toprule
            & PSNR$\uparrow$  & SSIM$\uparrow$ & LPIPS$\downarrow$ \\
            \midrule
            Ours & \textbf{32.63} & \textbf{0.928} & \textbf{0.083} \\
            Ours w/o LiDAR & 30.72 & 0.920 & 0.100 \\
            Ours w/o SfM & 29.97 & 0.911 & 0.106 \\
            Ours w/o Sky modeling & 31.12 & 0.921 & 0.100 \\
            \bottomrule
            \end{tabular}       
    }

    \label{tab:colmap_sky}

\end{table}

\begin{figure}[t]
    \centering
    \includegraphics[width=1\linewidth]{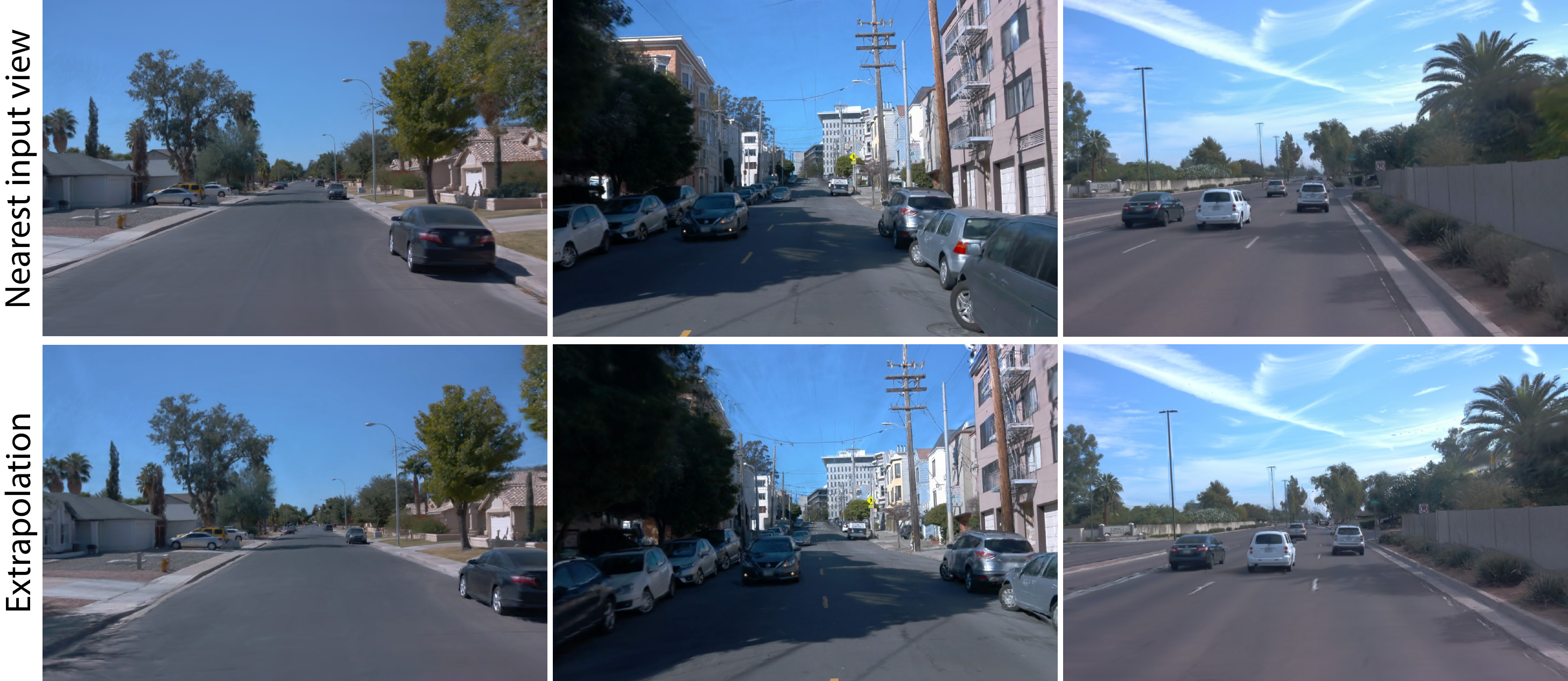}
    \caption{
        \textbf{Qualitative results of novel view synthesis with significant differences from the input frames.}
        In each scene we shift the camera by 2 meters.
    } 
\label{fig:extrapolation}
\end{figure}

\PAR{Extrapolation results.}
In Figure \ref{fig:extrapolation}, we show some qualitative results of novel view synthesis under the setting of lane changes on Waymo dataset.
Our method can produce high-quality results although the rendering viewpoint is far away from input sequence. 

\begin{figure*}[t]
    \centering
    \includegraphics[width=1\linewidth]{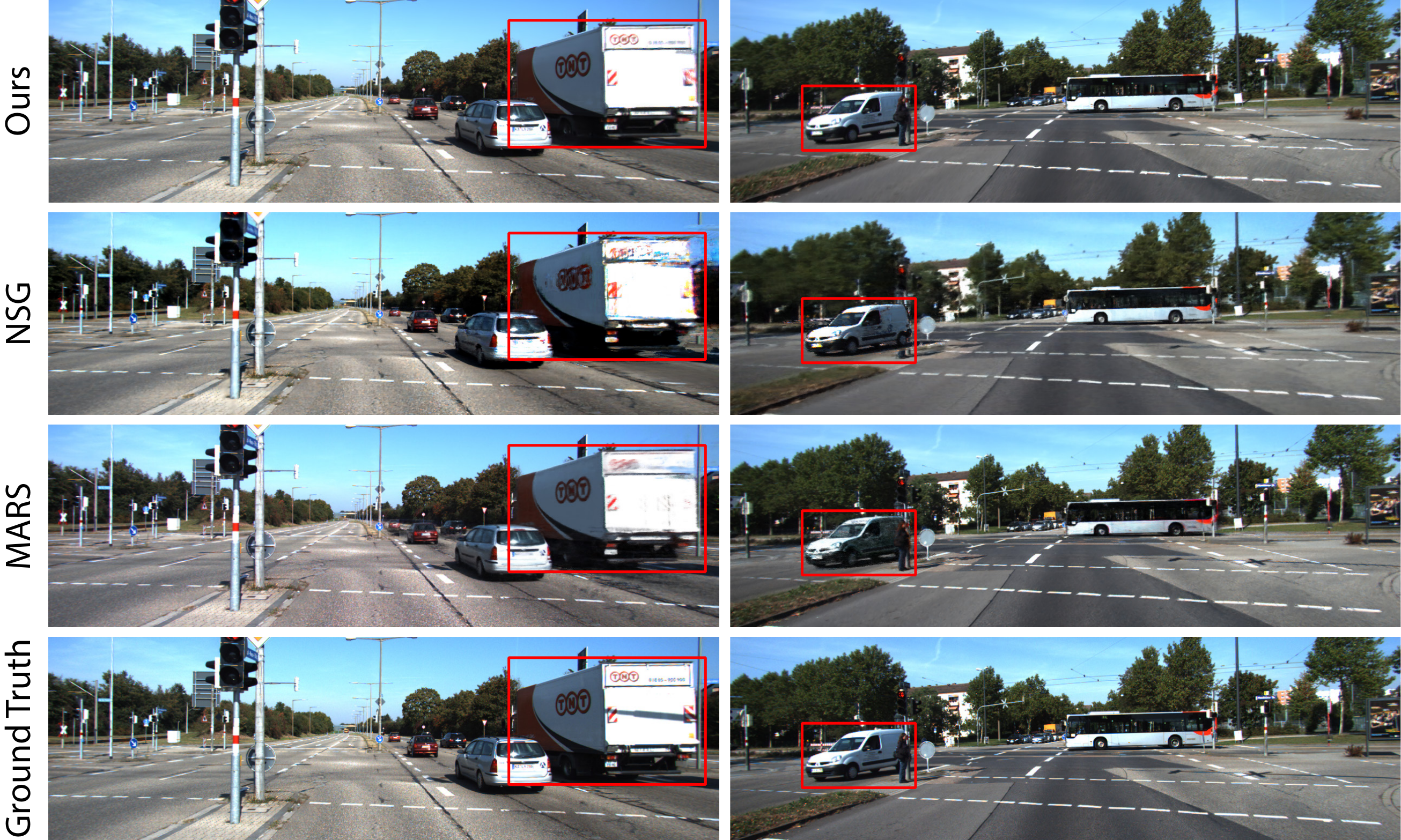}
    \caption{
        \textbf{Qualitative comparison results on the KITTI \cite{geiger2012we} dataset.} 
    }

    \label{fig:nvs_kitti}
\end{figure*}

\PAR{Qualitative results on the KITTI dataset.}
In Figure \ref{fig:nvs_kitti}, we show the comparison results with NSG \cite{Ost_2021_CVPR} and MARS \cite{wu2023mars} on the
KITTI \cite{geiger2012we} dataset.

\begin{figure*}[t]
    \centering
    \includegraphics[width=1.0\linewidth]{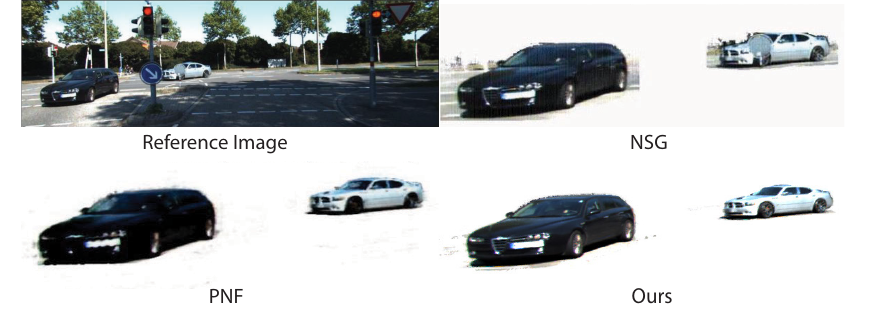}
    \caption{
        \textbf{Decomposition results on the KITTI \cite{geiger2012we} dataset.} 
    }

    \label{fig:decomposition_kitti}
\end{figure*}

\PAR{Decomposition results on the KITTI dataset.}
In Figure \ref{fig:decomposition_kitti}, we show the qualitative comparisons of decomposition with NSG \cite{Ost_2021_CVPR} and Panoptic Neural Fields \cite{KunduCVPR2022PNF} on the
KITTI \cite{geiger2012we} dataset.

\section{Limitations}
\methodnameblank also has some known limitations. 
1) Our method is limited to reconstructing rigid dynamic scenes, such as static streets with only moving vehicles, and cannot handle non-rigid dynamic objects like walking pedestrians. 
Future work could consider employing more complex dynamic scene modeling methods \cite{yang2023gs4d}, to address this issue.
2) the proposed method is dependent on the recall rate of off-the-shelf trackers. 
If some vehicles are missed, our pose optimization strategy cannot compensate for this. 
Obtaining continuous tracklets through methods like 2D tracking can alleviate the problem and modeling dynamic urban scenes without object tracklets remains an interesting problem.
3) \methodnameblank still requires per-scene optimization. We consider predicting generalizable 3D Gaussians in  
feed-forward manner as a future work.

\end{document}